\documentclass[10pt,twocolumn,letterpaper]{article}

\usepackage{iccv}
\usepackage{times}
\usepackage{epsfig}

\usepackage{graphicx}
\usepackage{amsmath}
\usepackage{amssymb}
\usepackage{booktabs}
\usepackage{multirow}
\usepackage{caption}
\usepackage{subcaption}
\usepackage{listings}
\usepackage{color}
\usepackage{xspace}
\usepackage{caption}
\definecolor{dkgreen}{rgb}{0,0.6,0}
\definecolor{gray}{rgb}{0.5,0.5,0.5}
\definecolor{mauve}{rgb}{0.58,0,0.82}

\graphicspath{ {images/} }
\usepackage[pagebackref=true,breaklinks=true,letterpaper=true,colorlinks,bookmarks=false]{hyperref}

\iccvfinalcopy 

\def\iccvPaperID{5074} 

\ificcvfinal\fi

\begin{document}

\title{Generative Action Description Prompts for Skeleton-based Action Recognition}

 \author{{Wangmeng Xiang\textsuperscript{1,}\textsuperscript{2} \qquad Chao Li\textsuperscript{2} \qquad Yuxuan Zhou\textsuperscript{2,3} \qquad Biao Wang\textsuperscript{2} \qquad Lei Zhang\textsuperscript{1}\thanks{Corresponding author} } \\
    	\textsuperscript{1}The Hong Kong Polytechnic University  \hspace{0.5em}
        \textsuperscript{2}DAMO Academy, Alibaba Group  \hspace{0.5em}
        \textsuperscript{3}Mannheim University \\
    	\tt\small \{wangmeng.xwm,lllcho.lc,wb.wangbiao\}@alibaba-inc.com, yuxuazho@mail.uni-mannheim.de \\ \tt\small cslzhang@comp.polyu.edu.hk
    }
\maketitle

\begin{abstract}
   Skeleton-based action recognition has recently received considerable attention. Current approaches to skeleton-based action recognition are typically formulated as one-hot classification tasks and do not fully exploit the semantic relations between actions. For example, ``make victory sign" and ``thumb up" are two actions of hand gestures, whose major difference lies in the movement of hands. This information is agnostic from the categorical one-hot encoding of action classes but could be unveiled from the action description. Therefore, utilizing action description in training could potentially benefit representation learning. In this work, we propose a $\textbf{G}$enerative $\textbf{A}$ction-description $\textbf{P}$rompts (GAP) approach for skeleton-based action recognition. More specifically, we employ a pre-trained large-scale language model as the knowledge engine to automatically generate text descriptions for body parts movements of actions, and propose a multi-modal training scheme by utilizing the text encoder to generate feature vectors for different body parts and supervise the skeleton encoder for action representation learning. Experiments show that our proposed GAP method achieves noticeable improvements over various baseline models without extra computation cost at inference. GAP achieves new state-of-the-arts on popular skeleton-based action recognition benchmarks, including NTU RGB+D, NTU RGB+D 120 and NW-UCLA. The source code is available at \url{https://github.com/MartinXM/GAP}.
\end{abstract}

\section{Introduction}

Action recognition has been an active research topic due to its wide range of applications in human-computer interaction, sports and health analysis, entertainment,~\etc ~In recent years, with the emergence of depth sensors, such as Kinect~\cite{zhang2012microsoft} and RealSense~\cite{realsense}, human body joints can be easily acquired. The action recognition approach utilizing body joints,~\ie, the so-called skeleton-based action recognition, has drawn a lot of attentions due to its computation efficiency and robustness to lighting conditions, viewpoint variations and background noise. 

Most of the previous methods in skeleton-based action recognition focus on modeling the relation of human joints, following a unimodal training scheme with a sequence of skeleton coordinates as inputs~\cite{zhang2017view,lee2017ensemble,si2019attention,du2015hierarchical,song2017end,cheng2020decoupling,shi2019two,ye2020dynamic,song2021constructing,xia2021multi,wang2021iip,plizzari2021spatial}. Inspired by the recent success of multi-modal training with image and language~\cite{radford2021learning,gpt3}, we investigate an interesting question: whether action language description could unveil the action relations and benefit skeleton-based action recognition? Regrettably, due to the absence of a large-scale dataset consisting of skeleton-text pairs, constructing such a dataset would require significant time and financial resources. Consequently, the training scheme outlined in~\cite{radford2021learning,jia2021scaling,yang2022unified} cannot be directly applied to skeleton-based action recognition. As a result, the development of novel multi-modal training paradigms is necessary to address this issue.


We propose to leverage the generative category-level human action description in the form of language prompts. The language definition of an action contains rich prior knowledge. For example, different actions focus on the movement of different body parts: ``make victory sign" and ``thumb up" describe the gesture of hands; ``arm circles" and ``tennis bat swing" describe the movement of arms; ``nod head" and ``shake head" are the motions of head; ``jump up" and ``side kick" rely on movements of foot and leg. Some actions describe the interaction of multiple body parts,~\eg, ``put on a hat" and ``put on a shoe" involve actions of hand and head, hand and foot, respectively. These prior knowledge about actions could provide fine-grained guidance for representation learning. In addition, to resolve the laborious work to collect human action prompts, we resort to pre-trained large language model (LLM), e.g. GPT-3~\cite{gpt3} for efficient automatic prompts generation.

	\begin{figure*}[h]
		
		\begin{center}
			\includegraphics[width=0.95\textwidth]{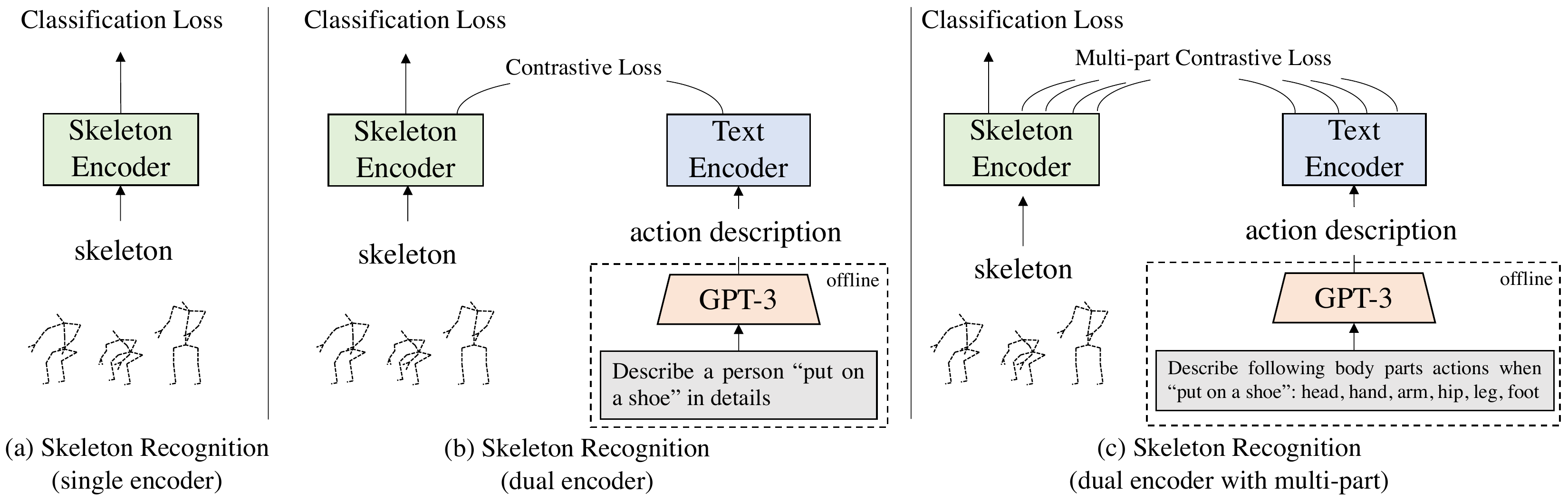}		
			\caption{Comparison of our proposed Generative Action-description Prompts (GAP) framework (dual encoder) with other skeleton recognition methods (single encoder). Besides classification loss, our proposed method contains additional contrastive loss. Notice that text encoder is only used at the training stage and GPT-3 is applied for offline action description generation. For every given action query, GPT-3 generates text description of actions with prompt templates, the action description is then employed for multi-modal training.}
			\vspace{-8mm}
			\label{fig:framework}
		\end{center}
		
	\end{figure*}

In specific, we develop a new training paradigm, which employs generative action prompts for skeleton-based action recognition. We take advantages of the GPT-3~\cite{gpt3} as our knowledge engine to generate meaningful text descriptions for actions. With elaborately designed text prompts, detailed text descriptions for the whole action and each body part can be produced. In Figure~\ref{fig:framework}, we compare our proposed frameworks (b) and (c) with traditional single encoder skeleton-based action recognition framework (a). In our framework, a multi-modal training scheme is developed, which contains a skeleton encoder and a text encoder. The skeleton encoder takes skeleton coordinates as inputs and generates both part feature vectors and global feature representations. The text encoder transforms global action description or body part descriptions into text features for the whole action or each body part. A multi-part contrastive loss (single contrastive loss for (b)) is used to align the text part features and skeleton part features, and the cross-entropy loss is applied on the global features. 


Our contributions are summarized as follow:
\begin{itemize}
    \item[-] As far as we known, this is the first work to use generative prompts for skeleton-based action recognition, which applies a LLM as the knowledge engine and elaborately employs text prompts to generate detailed text descriptions of the whole action and body parts movements for different actions automatically.
    \item[-] We propose a new multi-modal training paradigm that utilizes generative action prompts to guide skeleton-based action recognition, which enhances the representation by using knowledge about actions and human body parts. It could improve the model performance without bringing any computation cost at inference.
    \item[-] With the proposed training paradigm, we achieve state-of-the-art performance on several popular skeleton-based action recognition benchmarks, including NTU RGB+D, NTU RGB+D 120 and NW-UCLA.
\end{itemize}

 \begin{figure*}[h]
	
	\begin{center}
		\includegraphics[width=0.95\textwidth]{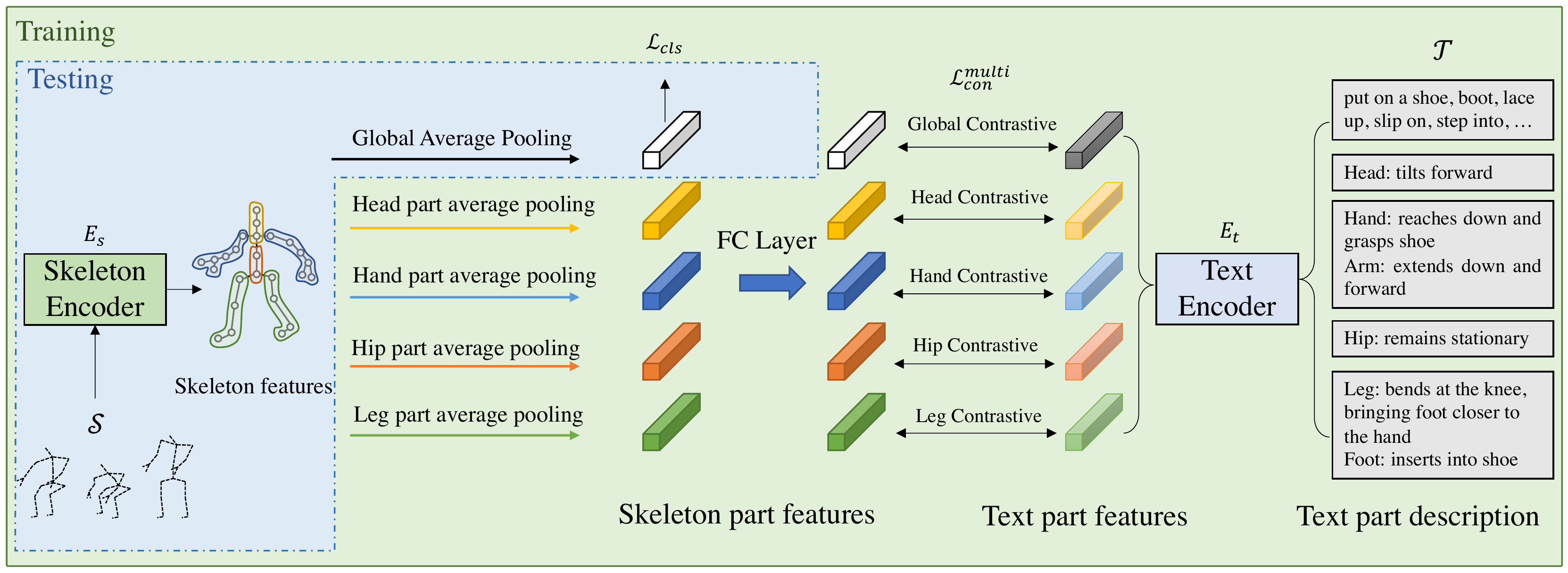}		
		\caption{ Overall framework of Generative Action-description Prompts (GAP) with multi-part contrastive loss. During training, the skeleton encoder is trained with both cross-entropy loss and multi-part contrastive loss. We use GPT-3 generated action description as input for text encoder to generate human part features. The part features are then aligned with skeleton encoder part features with multi-part contrastive loss. During testing, only global feature from skeleton encoder is used for classification, and text encoder is neglected. }
		\vspace{-8mm}
		\label{fig:multiframework}
	\end{center}
	
\end{figure*}

\section{Related work}

\subsection{Skeleton-based Action Recognition}

In recent years, various methods have been proposed for skeleton-based action recognition by designing efficient and effective model architecture. RNNs were applied to handle the sequence of human joints in ~\cite{du2015hierarchical,song2017end,zhang2017view}. HBRNN~\cite{du2015hierarchical} employed an end-to-end hierarchical RNN to model long-term contextual information of temporal skeleton sequences. VA-LSTM~\cite{zhang2017view} designed a view adaptive RNN, which enables the network to adapt to the most suitable observation viewpoints from end to end. Inspired by the success of CNN in image tasks, CNN-based methods~\cite{zhang2019view, xu2021topology} have been utilized to model joints relations. A pure CNN architecture named Topology-aware CNN (TA-CNN) is proposed in~\cite{xu2021topology}. As human joints can be naturally presented as graph nodes and joint connections can be described by adjacent matrix, GCN-based methods~\cite{yan2018spatial,cheng2020decoupling,shi2019two,chen2021channel,song2021constructing} have drawn a lot of attentions. For example, ST-GCN~\cite{yan2018spatial} applied spatial-temporal GCN to model human joints relations in both spatial and temporal dimension. CTR-GCN~\cite{chen2021channel} proposed a channel-wise graph convolution for fine-grained relation modeling. Info-GCN~\cite{Chi_2022_CVPR} adopt an information bottleneck in GCN. With the recent popularity of vision transformer~\cite{dosovitskiy2021an}, transformer-based methods~\cite{plizzari2021spatial, shi2020decoupled, wang2021iip} have also been investigated for skeleton data. All the previous methods adopt a unimodal training scheme. As far as we known, our work is the first to apply a multi-modal training scheme for skeleton-based action recognition.

\subsection{Human Part Prior}

Human part prior for skeleton-based action recognition has been used by designing special model architectures in previous works~\cite{thakkar2018part,song2020stronger,wang2021iip,huang2020part}. PB-GCN~\cite{thakkar2018part} divided the skeleton graph into four subgraphs and learned a recognition model using a part-based graph convolutional network. PA-ResGCN~\cite{song2020stronger} calculated attention weights for human body parts to improve the discriminative capability of the features. PL-GCN~\cite{huang2020part} proposed a part-level graph convolutional network to automatically learn the part partition strategy. IIP-transformer~\cite{wang2021iip} applied transformer to learn inter-part and intra-part relations. Comparing to previous methods, we directly use part language description to guide representation learning during training with a multi-part contrastive loss. We do not design any complicated part modeling module and thus do not introduce extra computation cost at inference.

\subsection{Multi-modal Representation Learning}

Multimodal representation learning methods, such as CLIP~\cite{radford2021learning} and ALIGN~\cite{jia2021scaling}, have shown that vision-language co-training can learn powerful representation for downstream tasks such as zero-shot learning, image captioning, text-image retrieval,~\etc. UniCL~\cite{yang2022unified} uses a unified contrastive learning method that regards image-label as image-text-label data to learn the generic visual-semantic space. However, these methods require a large-scale image-text paired dataset for training. ActionCLIP~\cite{wang2021actionclip} follows the training scheme of CLIP for video action recognition. A pre-trained CLIP model is used and transformer layers are added for temporal modeling of video data. As for action description, label names are directly used as text prompts with prefix and suffix that do not contain much semantic meanings,~\eg,\textit{``A video of [action name]"}, \textit{``Human action of [action name]"},~\etc~In contrast, we use a LLM (GPT-3), as knowledge engine to generate descriptions of human body movements in actions, which provide fine-grained guidance for representation learning. In addition, we employ multi-part contrastive loss on body parts to learn a fine-grained skeleton representation. Prompt Learning (PL)~\cite{zhou2022coop,zhou2022cocoop,ju2022prompting} approaches aim to tackle the challenges posed by zero-shot and few-shot learning by through the incorporation of learnable prompt vectors. While PL has demonstrated promising results, the interpretability of the learned prompt vectors remains a challenge. Recently,~\cite{SachitMenon2022VisualCV} applies LLM for generating descriptions for zero-shot image classification. STALE~\cite{nag2022zero} applies parallel classification and localization/classification architecture for zero-shot action detection. MotionCLIP~\cite{tevet2022motionclip} is proposed to align action latent space with CLIP latent space for 3D human action generation. ActionGPT~\cite{Action-GPT} uses LLM to generate detailed action description for action generation. Our research is conducted concurrently and independently. All these methods require a text encoder during inference, whereas our proposed framework only imposes overheads during the training phase, without adding any computational or memory costs during testing.


\vspace{-2mm}

\section{Methods}

In this section, we present in detail the proposed \textbf{G}enerative \textbf{A}ction-description \textbf{P}rompts (GAP) framework. GAP aims to enhance skeleton representation learning with automatically generated action descriptions and it can be embedded into the existing backbone networks. Therefore, GAP can be coupled with various skeleton and language encoders. In the following sections, we first overview the GAP framework, then introduce the skeleton encoder, text encoder and the main components of GAP in detail.

\subsection{Generative Action Prompts Framework} 

The comprehensive framework of our GAP approach is presented in Figure~\ref{fig:multiframework}. It is composed of a \textbf{skeleton encoder} $E_s$ and a \textbf{text encoder} $E_t$, for generating skeleton features and text features, respectively. The training loss can be presented as:
\begin{equation}
	\begin{aligned}
 \mathcal{L}_{total} = \mathcal{L}_{cls}(E_s(\mathcal{S})) + \lambda \mathcal{L}^{multi}_{con}(E_s(\mathcal{S}), E_t(\mathcal{T})),
	\end{aligned}
	\label{eq:train}
\end{equation}
\noindent where, $\mathcal{L}_{cls}$ is cross-entropy classification loss, $\mathcal{L}^{multi}_{con}$ is multi-part contrastive loss. Skeleton input $\mathcal{S} \in \mathbb{R}^{B \times 3\times N\times T}$, $B$ is the batch size, 3 is the coordinate number, $N$ and $T$ are joint number and sequence length, respectively. $\lambda$ is a learnable trade-off parameter. $\mathcal{T}$ is LLM generated text descriptions. 

During training, the $E_s$ is trained with cross-entropy loss and multi-part contrastive loss with part text descriptions as additional guidance. The global skeleton feature is generated by performing average pooling of all joint nodes and the part skeleton features are generated by aggregating the features of various groups of nodes using average pooling. The skeleton part features are mapped by fully connected layer (FC Layer) to keep the same feature dimension as text features. The text part descriptions are generated by LLM offline, and encoded by $E_t$ during training for producing text part features. At the testing stage, we directly use global features of skeleton encoder for action probability prediction. Therefore, our GAP framework does not bring additional memory or computation cost at inference comparing to previous skeleton encoder only method.





\subsection{Skeleton Encoder}

Graph Convolution Network (GCN) is prevailing for skeleton action recognition due to its efficiency and strong performance. Therefore, we adopt GCN as the backbone network in our GAP framework. Our skeleton encoder consists of multiple GC-MTC blocks, while each block contains a graph convolution (GC) layer and a multiscale temporal convolution (MTC) module.

 \textbf{Graph Convolution.} The human skeleton can be represented as a graph $G = \{V,\mathcal{E}\}$, where $V$ is the set of human joints with $|V| = N$, and $\mathcal{E}$ is the set of edges. Denote by $\mathbf{H}^{l} \in \mathbb{R}^{N\times F}$ the features of human joints at layer $l$ with feature dimension $F$. The graph convolution can be formulated as follows:
\begin{equation}
	\begin{aligned}
    \mathbf{H}^{l+1} = \sigma (\mathbf{D}^{-\frac{1}{2}} \mathbf{A}\mathbf{D}^{-\frac{1}{2}} \mathbf{H}^{l} \mathbf{W}^{l}),
\end{aligned}
\label{eq:gcn}
\end{equation}

\noindent where $\mathbf{D} \in \mathbb{R}^{N\times N}$ 
is the degree matrix, $\mathbf{A}$ is the adjacency matrix representing joints connections, $\mathbf{W}^{l}$ is the learnable parameter of the $l$-th layer and $\sigma$ is the activation function. 

 \textbf{Multiscale Temporal Modeling.} To model the action at different temporal speed, we utilize the multiscale temporal convolution module in~\cite{liu2020disentangling,chen2021channel} for temporal modeling. The module comprises four distinct branches, each of which incorporates a $1 \times 1$ convolution to decrease channel dimensionality. There are two temporal convolutions branches with varing dilations (1 and 2) and one MaxPool branch. The fourth branch only contains $1\times1$ convolution. The outputs of the four branches are concatenated to produce the final result.
 

\textbf{Skeleton Classification.} The skeleton-based action recognition methods map human skeleton data to one-hot encoding of action labels, which are trained with a cross-entropy loss:
\begin{equation}
	\begin{aligned}
    \mathcal{L}_{cls} = -y\log p_{\theta}(x),
\end{aligned}
\label{eq:cls}
\end{equation}
\noindent where $y$ is the one-hot ground-truth action label, $x$ is the global skeleton feature and $p_{\theta}(x)$ is the predicted probability distribution.

\subsection{Text Encoder}
Considering the recent success of Transformer models in NLP, we employ a pre-trained transformer-based language model as our text encoder $E_t$, such as BERT~\cite{devlin2018bert} or CLIP-text-encoder~\cite{radford2021learning}. The input is in the form of text and undergoes a standard tokenization process. Subsequently, the features are processed through a series of transformer blocks. The final output is a feature vector that represents the text description. For different human part, we use various part descriptions as text encoder's input.


\subsection{Action Description Learning}

\textbf{Skeleton-language Contrastive Learning.} Comparing to the one-hot label supervision for skeleton classification, skeleton-language contrastive learning employs the supervision from natural language. It has a dual-encoder design with a skeleton encoder $E_s$ and a text encoder $E_t$, which encode skeleton data and action descriptions, respectively. The dual-encoders are jointly optimized by contrasting skeleton-text pairs in two directions within the batch:
\begin{equation}
	\begin{aligned}
    p_{i}^{s2t}(\mathbf{s_i}) = \frac{\exp(\text{sim}(\mathbf{s_i},\mathbf{t_{i}})/\tau)}{\sum ^{B} _{j=1} \exp(\text{sim}(\mathbf{s_i},\mathbf{t_{j}})/\tau)}, \\
    p_{i}^{t2s}(\mathbf{t_i}) = \frac{\exp(\text{sim}(\mathbf{t_{i}},\mathbf{s_{i}})/\tau)}{\sum ^{B} _{j=1} \exp(\text{sim}(\mathbf{t_i},\mathbf{s_{j}})/\tau)},
\end{aligned}
\label{eq:prob}
\end{equation}
\noindent where $\mathbf{s}$, $\mathbf{t}$ are encoded features of skeleton and text, $\text{sim}(\mathbf{s},\mathbf{t})$ is the cosine similarity, $\tau$ is the temperature parameter and $B$ is the batch size. Unlike image-text pairs in CLIP, which are one-to-one mappings, in our setting, there could be more than one positive matching and actions of different categories forming negative pairs. Therefore, instead of using cross-entropy loss, we use KL divergence as the skeleton-text contrastive loss: 
\begin{equation}
	\begin{aligned}
    \mathcal{L}_{con} = \frac{1}{2} \mathbf{E}_{\mathbf{s},\mathbf{t} \sim \mathcal{D}}  [KL(p^{s2t}(\textbf{s}), y^{s2t}) + KL(p^{t2s}(\textbf{t}), y^{t2s})],
\end{aligned}
\label{eq:kl}
\end{equation}
\noindent where $\mathcal{D}$ is the entire dataset, $y^{s2t}$ and $y^{t2s}$ are ground-truth similarity scores, which have a probability of 0 for negative pairs and a probability of 1 for positive.

	\begin{figure}[t]
		
		\begin{center}
			\includegraphics[width=0.44\textwidth]{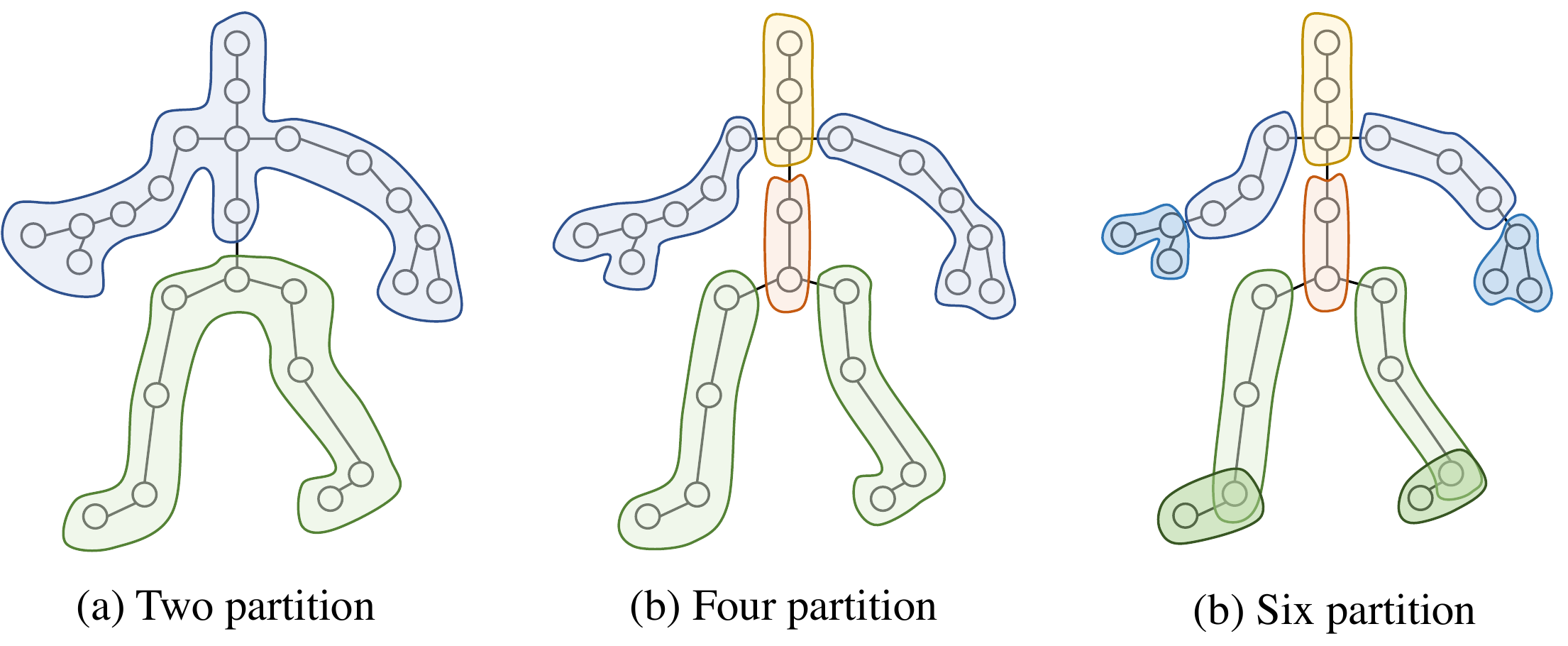}		
			\caption{Different part partition strategies. (a) Two parts: upper and lower body. (b) Four parts: head, hand-arm, hip, leg-foot. (c) Six parts: head, arm, hand, hip, leg, foot.}
			\vspace{-5mm}
			\label{fig:part}
		\end{center}
		
	\end{figure}

\textbf{Multi-part Contrastive Learning.} Considering the prior of human body parts, skeleton can be divided into multiple groups. We illustrate this framework in Figure~\ref{fig:framework}(c). We apply contrastive loss on different parts features as well as global feature, and propose a multi-part contrastive loss. The part feature could be obtained with part pooling, where joint features within the same group are aggregated to generate part representation. More specifically, we choose the features before the final classification layer for part feature pooling. In Figure~\ref{fig:part}, we show different part partition strategies. For two parts partition, the whole body is divided into upper and lower groups. For four parts partition, the body is divided into four groups: head, hand-arm, hip, leg-foot. For six parts partition, head, hand, arm, hip, leg, foot are grouped separately. The loss function of multi-part contrastive loss can be represented as follows:

\begin{equation}
	\begin{aligned}
    \mathcal{L}^{multi}_{con} = \frac{1}{K} \sum ^{K}_{k=1} \mathcal{L}^{k}_{con},
\end{aligned}
\label{eq:total}
\end{equation}

\noindent where $K$ is the total part number.

	\begin{figure}[t]
		
		\begin{center}
			\includegraphics[width=0.45\textwidth]{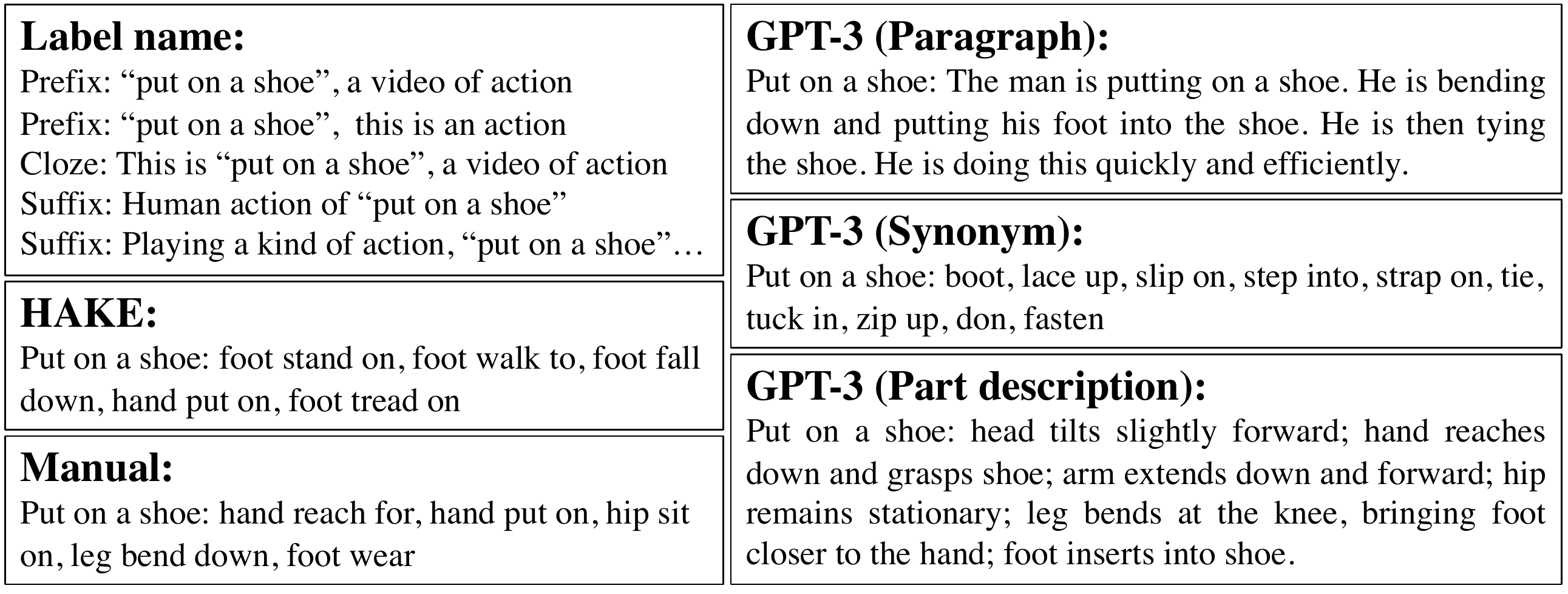}		
			\caption{Text description generated by different methods.}
			\vspace{-8mm}
			\label{fig:prompt}
		\end{center}
		
	\end{figure}

		\begin{figure}[t]
		
		\begin{center}
			\includegraphics[width=0.4\textwidth]{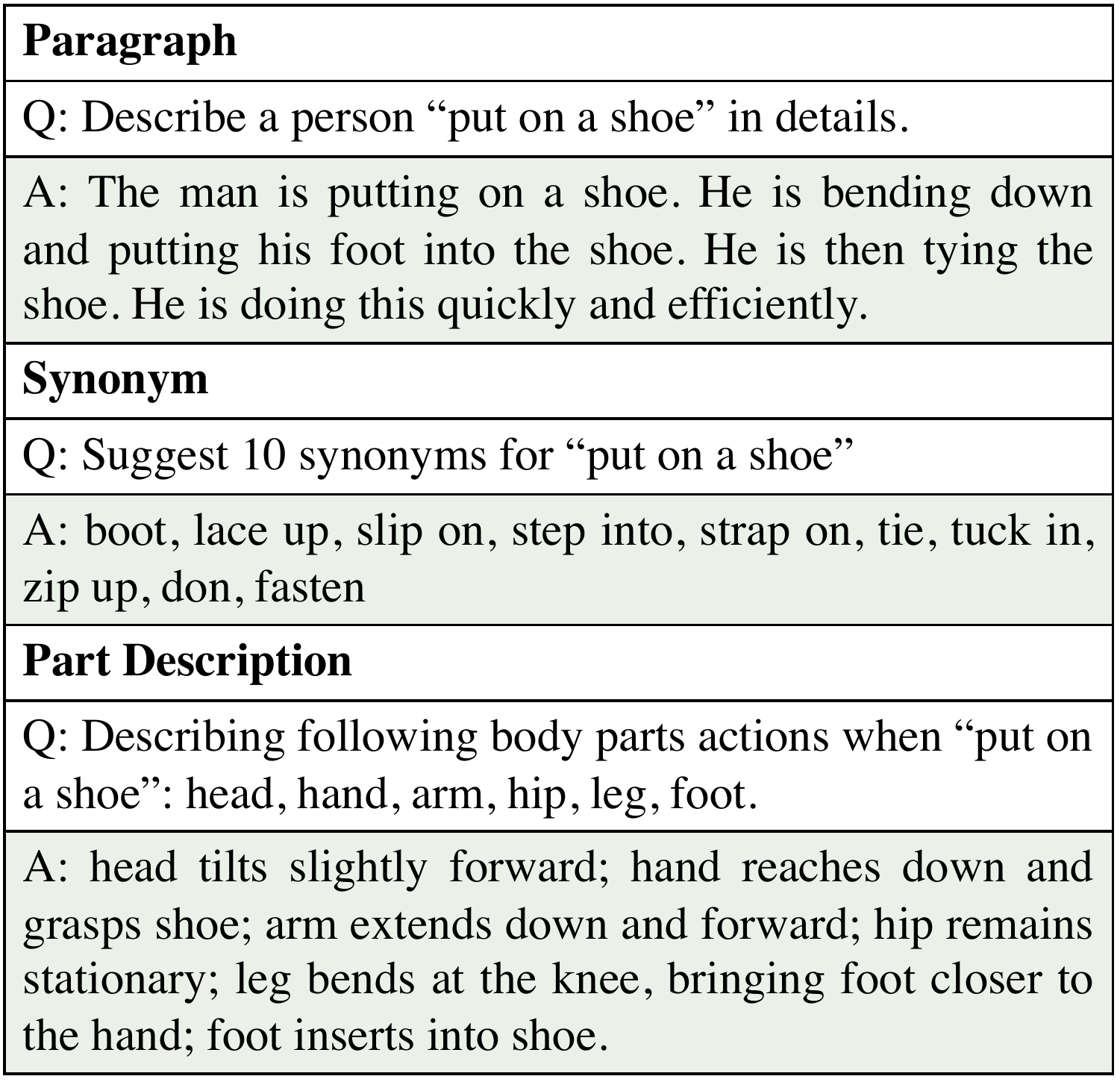}		
			\caption{Text description generated from different prompt inputs by GPT-3.}
			\label{fig:prompt_gpt3}
		\end{center}
			\vspace{-8mm}
	\end{figure}

\subsection{Action Description Generation}

The action description $\mathcal{T}$ for text encoder plays a vital role in GAP. Here, we explore several different description generation methods. Figure~\ref{fig:prompt} illustrates the text descriptions of action ``put on a shoe" by different methods.

 \textbf{Label Name.} One straight-forward approach is to directly  use the label name. Many methods~\cite{wang2021actionclip} use this kind of text descriptions with prefix and suffix such as ``Human action of [action]", ``[action], a video of action",~\etc~Though these prompts could boost the performance for zero-shot and few-shot problems, in our case of supervised learning, this approach does not bring significant performance improvement (as shown in our ablation studies) since these prompts do not contain discriminative semantic information about actions.

 \textbf{HAKE Part State.} The HAKE~\cite{li2023hake} dataset contains annotated part states of human-object interactions. For each sample, six body part movements (head, hand, arm, hip, leg, foot) are manually annotated, with 93 part states in total. In order to avoid laborious annotation for each sample, we apply an automatic pipeline which contains two steps: 1) generate text features for both label name and HAKE part states with a pre-trained transformer text encoder; 2) generate text description by finding the $K$ nearest neighbors of action label name in HAKE part state feature space. Those HAKE part states that are closest to the action label name are selected for action description. We then use this generated part description for GAP.

 \textbf{Manual Description.} We ask annotators to write down the description of body part movements following the temporal order of the action. The descriptions consist of the predefined atomic movements. The annotators are asked to focus on the most distinguished parts' motions.

 \textbf{Large-language Model.} We use the large-scale language model (\eg, GPT-3) to generate text descriptions. We design text prompts so that it can generate our desired action descriptions. Text descriptions are generated in three ways. a) \textit{paragraph}: a full paragraph that can describe the action in detail; b) \textit{synonym}: we collect 10 synonyms of action labels; c) \textit{part description}: we collect descriptions of different body parts for each action. The body partition strategies follow Figure~\ref{fig:part} in previous section. We take ``put on a shoe" as an example and present the prompts used for generating different descriptions in Figure~\ref{fig:prompt_gpt3}.

\begin{table*}[h]
\begin{subtable}[c]{0.35\textwidth}
   \subcaption{Partition strategies}
  \centering
  \begin{tabular}{l@{}c@{}}
    \toprule
    Partition Strategy & Acc(\%) \\
    \midrule
    None    & 84.6 \\
    Global  & 85.2 \\
    Upper, Lower  & 85.3 \\
    Head, Hand, Hip, Leg  & \textbf{85.4} \\
    Head, Hand, Arm, Hip, Leg, Foot  & \textbf{85.4} \\
    \bottomrule
  \end{tabular}
  \label{tab:partition}
    \end{subtable}
  \begin{subtable}[c]{0.34\textwidth}
   \subcaption{Text prompt type}
  \centering
  \begin{tabular}{lc}
    \toprule
    Prompt type & Acc(\%) \\
    \midrule
    None & 84.6 \\
    Label name & 84.8$(\uparrow0.2)$ \\ 
    Synonym/Paragraph  & 85.2$(\uparrow0.6)$ \\
    Body parts  & 85.4$(\uparrow0.8)$ \\
    Synonym+Body parts & \textbf{85.5$(\uparrow0.9)$} \\

    \bottomrule
  \end{tabular}
  \label{tab:prompt}
    \end{subtable}
    \begin{subtable}[c]{0.33\textwidth}
   \subcaption{Text encoders}
  \centering
  \begin{tabular}{ccc}
    \toprule
    Text encoder & pretrain & Acc(\%) \\
    \midrule
        XFMR-32 & img/text  & \textbf{85.2} \\
        XFMR-16 & img/text & 85.1 \\
        XFMR-14 & img/text & \textbf{85.2} \\
        BERT & text  & \textbf{85.2} \\
    \bottomrule
  \end{tabular}
  \label{tab:encoder}
    \end{subtable}
\begin{subtable}[c]{0.4\textwidth}
   \subcaption{Effect of GAP on different skeleton encoders}
  \centering
  \begin{tabular}{l@{}c@{}c}
    \toprule
    \multirow{2}*{Backbone} & \multicolumn{2}{c}{Acc(\%)} \\
    &  w./o. GAP &  w. GAP  \\
    \midrule
        ST-GCN~\cite{yan2018spatial}  & 82.6 & 83.8$(\uparrow1.2)$\\
        CTR-baseline & 83.7 & 84.6$(\uparrow0.9)$\\
        CTR-GCN (single scale) & 84.6 & \textbf{85.5}$(\uparrow0.9)$  \\
        CTR-GCN (multi scale)~\cite{chen2021channel}  & 84.9 & \textbf{85.5}$(\uparrow0.6)$  \\

    \bottomrule
  \end{tabular}
  \label{tab:backbone}
    \end{subtable}
\begin{subtable}[c]{0.31\textwidth}
   \subcaption{Description methods}
  \centering
  \begin{tabular}{lc}
    \toprule
    Methods & Acc(\%) \\
    \midrule
    Part CLS Baseline & 84.2 \\
    Manual description  & 85.2 \\
    HAKE part state& 85.3 \\
    GPT-3 generated  & \textbf{85.5} \\
    \bottomrule
  \end{tabular}
  \label{tab:source}
    \end{subtable}
 \begin{subtable}[c]{0.33\textwidth}
	   \subcaption{Comparison with Prompt Learning}
	  \centering
      \begin{tabular}{@{}cc@{}cc}
\toprule
	Methods &Prompt & TE & Acc  \\
\midrule
	Baseline & Fixed & Tuned &  84.8  \\
	\multirow{2}{*}{PL\cite{zhou2022coop}} & Learned & Fixed &  85.1  \\
	& Learned & Tuned &  85.2  \\
	GAP   & Generated  &  Tuned   &  \textbf{85.5}   \\
\bottomrule
\end{tabular}
\label{tab:promptlearning}
 \end{subtable}

     \caption{Ablation study of different components of GAP on NTU120, including partition strategy, text prompt type, text encoder, skeleton encoder, prompt methods. The Acc represents action recognition accuracy, and TE represents text encoder.}
     \label{tab:ablation}
    \vspace{-3mm}

\end{table*}

%

\section{Experiments}





\subsection{Datasets}

\textbf{NTU RGB+D} 
\cite{shahroudy2016ntu} is a widely used dataset for skeleton-based human action recognition. It contains
56,880 skeletal action sequences. There are two benchmarks for evaluation, including Cross-Subject (X-Sub) and Cross-View (X-View) settings. For X-Sub, the training and test sets come from two disjoint sets, each having 20 subjects. For X-View, the training set contains 37,920 samples captured by camera views 2 and 3, and the test set includes 18,960 sequences captured by camera view 1. 

\textbf{NTU RGB+D 120} \cite{liu2019ntu} is an extension of NTU RGB+D dataset with 57,367 additional skeleton sequences over 60 additional action classes. There are 120 action classes in total. Two benchmark evaluations were suggested by the authors, including Cross-Subject (X-Sub) and Cross-Setup (X-Setup) settings. 

\textbf{NW-UCLA} \cite{wang2014cross} dataset is
recorded by three Kinect V1 sensors from different viewpoints. The skeleton contains 20 joints and 19 bone connections. It includes 1,494 video sequences of 10 action categories.

\subsection{Implementation Details}

For NTU RGB+D and NTU RGB+D 120, each sample is resized to 64 frames, and we adopt the code of \cite{zhang2020semantics,Chi_2022_CVPR} for data pre-processing. For NW-UCLA, we follow the data pre-processing procedures in~\cite{cheng2020skeleton,chen2021channel,Chi_2022_CVPR}. We use CTR-GCN with single-scale temporal convolution for our ablation study, considering its good balance between performance and efficiency. For ablation study with ST-GCN backbone, please refer to supplementary material. When comparing with other methods, we adopt CTR-GCN with multiscale temporal convolution since it produces the best results. For text encoder, we use the pretrained text transformer model from CLIP or BERT and finetune its parameters during training. The temperature of contrastive loss is set to $0.1$. As for the non-deterministic of action descriptions generated by GPT-3, we effectively employed generated results through sampling in the course of training. For example, in the context of our synonyms scenario, we generate numerous synonyms and select them randomly for use in training.

For NTU RGB+D and NTU RGB+D 120, we train the model for a total number of 110 epochs with batch size 200. We use a warm-up strategy for the first 5 epochs. The initial learning rate is set to 0.1 and reduced by a factor of 10 at 90 and 100 epochs, the weight decay is set to 5e-4 following the strategy in~\cite{Chi_2022_CVPR}. For NW-UCLA, the batchsize, epochs, learning rate,weight decay, reduced step, warm-up epochs are set to 64, 110, 0.2, 4e-4, [90,100], 5, respectively.

%
%
%
%
%
%
%
%
%
%
%

\subsection{Ablation Study}

In this section, we conduct experiments to evaluate the influences of different components. The experiments are conducted on NTU120 RGB+D with joint modality and X-Sub setting. For more ablation studies please refer to \textbf{supplementary materials}.


\textbf{Partition Strategies.} We test different body partition strategies for GAP and the results are shown in Table~\ref{tab:partition}. `Global' represents using a global description of actions with a single contrastive loss, and it improves over the baseline by 0.6\%. Using more parts and multi-part contrastive loss could steadily increase the performance, and it saturates at 85.4\% when using 4 parts.

\textbf{Influences of Text Prompt.} The text prompt design has a large impact on the model performance. We show the influences of different text prompts in Table~\ref{tab:prompt}. By directly using label name (with prefix or suffix) as the text prompt in GAP, the model only slightly outperforms (0.2\%) the baseline model without text encoder, as this does not bring extra information for training. Utilizing a synonym list for label name or a global description paragraph could largely improve the performance (0.6\%) over baseline, as it enriches the semantic meanings of each action class. Using part description prompts leads to strong performance with 0.8\% improvement. The best performance is achieved by combining synonym of label name and body part description for prompts, resulting in 85.5\% accuracy.


\textbf{Influences of Text Encoder.} In Table~\ref{tab:encoder}, we show the influences of text encoders. We found that both XFMR (text encoder from CLIP~\cite{radford2021learning}) and BERT all achieve good performance, indicating that skeleton encoder could benefit from text encoder with different pre-training sources (image-language or pure language). We use XFMR-32 as our default text encoder considering its good balance between efficiency and accuracy.

\textbf{Effect of GAP on Different Skeleton Encoders.} Our proposed GAP is decoupled from the network architecture and could be employed to improve different skeleton encoders. In Table~\ref{tab:backbone}, we show experimental results of applying GAP to ST-GCN~\cite{yan2018spatial}, CTR-baseline and CTR-GCN~\cite{chen2021channel}. GAP brings consistent improvements ($0.6$-$1.2\%$) without extra computation cost at inference, demonstrating the effectiveness and generalization ability of GAP.

\textbf{Comparison of Description Methods.} We compare several different methods of obtaining text prompts for text encoders in Table~\ref{tab:source}, including: Manual description; HAKE part state; Generating text prompts with GPT-3. For manual descriptions and HAKE results, we use them as global description for GAP. Among these methods, GPT-3 could provide very detailed description of human parts by using an elaborately designed text prompt, and the generated part text description achieves the best performance. We also implement a part pooling classification baseline for reference, which applies a classification head for every pooled part feature. This baseline does not work well as the part feature may not be sufficient to predict the action classes. 


\textbf{Comparing with prompt learning methods.} In Table~\ref{tab:promptlearning}, we compare GAP with PL methods that make prompts learnable parameters. PL outperforms baseline with both Text Encoder (TE)'s parameter fixed or tuned. GAP further outperforms PL by $0.3\%$, which indicates the effectiveness generated prompts and the multi-part paradigm.

\textbf{Influences of $\lambda$ Selection.} To study the influences of trade-off parameter $\lambda$ in Eq.~\ref{eq:train}, we search the value of $\lambda$ in $\{1.0, 0.8, 0.5, 0.2\}$ with 5-fold cross-validation. The performance of models are $85.4\%$, $85.5\%$, $85.3\%$ and $85.2\%$, respectively. We found that $\lambda = 0.8$ achieves the best performance; therefore, we utilize it as our default $\lambda$ value and employ it for all the experiments on different benchmarks.

\begin{table}[ht]
  \centering
  \resizebox{0.4\textwidth}{24mm}{
  \begin{tabular}{lc@{}c@{}}
    \toprule
        \multirow{2}*{Methods} &  \multirow{2}*{Mode} & \multicolumn{1}{c}{NW-UCLA} \\
        & & Top-1 (\%)  \\
    \midrule
   
        Ensemble TS-LSTM \cite{lee2017ensemble} & 2 ensemble & 89.2 \\
        2S-AGC-LSTM \cite{si2019attention}& 2 ensemble & 93.3 \\
      4S-Shift-GCN \cite{cheng2020skeleton} & 4 ensemble & 94.6 \\
    DC-GCN+ADG \cite{cheng2020decoupling} & 4 ensemble & 95.3 \\ 
          TA-CNN \cite{xu2021topology} & 4 ensemble & 96.1 \\
        CTR-GCN \cite{chen2021channel} & 4 ensemble & 96.5 \\
        Info-GCN \cite{Chi_2022_CVPR} & 4 ensemble & 96.6 \\ 
        \hline
        \multirow{3}*{\textbf{Ours}} & Joint/Joint-M & 94.0/93.5 \\
         & Bone/Bone-M & 95.3/91.2 \\
         & 4 ensemble & \textbf{97.2} \\
    \bottomrule
  \end{tabular}
  }
    \caption{Action classiﬁcation performance on the NW-UCLA dataset.}
  \label{tab:ucla}
     \vspace{-5mm}
\end{table}


\begin{table}[ht]
  \centering
      \resizebox{0.45\textwidth}{42mm}{

  \begin{tabular}{l@{}c@{}c@{}c@{}}
    \toprule
\multirow{2}*{Methods} & \multirow{2}*{Mode} & \multicolumn{2}{c}{NTU-RGB+D} \\
        &    &  X-Sub(\%) & X-View(\%)   \\
    \midrule

    VA-LSTM~\cite{zhang2017view} & 2 ensemble & 79.4 & 87.6 \\
    HCN~\cite{li2018co} & 2 ensemble & 86.5 & 91.1 \\
    2S-AGCN~\cite{shi2019two} & 2 ensemble & 88.5 & 95.1 \\

    SGN~\cite{zhang2020semantics} & 2 ensemble & 89.0 & 94.5 \\
    2S-AGC-LSTM~\cite{si2019attention} & 2 ensemble & 89.2 & 95.0 \\ 

    ST-TR (Plizzari et al. 2021) & 4 ensemble & 89.9 & 96.1    \\
    TA-CNN~\cite{xu2021topology} & 4 ensemble & 90.4 & 94.8 \\

    4S-Shift-GCN~\cite{cheng2020skeleton} & 4 ensemble & 90.7 & 96.5 \\

    DC-GCN+ADG~\cite{cheng2020decoupling} & 4 ensemble & 90.8 & 96.6 \\ 
    PA-ResGCN-B19~\cite{song2020stronger} & 4 ensemble & 90.9  & 96.0 \\
    Dynamic GCN~\cite{ye2020dynamic} & 4 ensemble & 91.5 & 96.0 \\
    MS-G3D~\cite{liu2020disentangling} & 2 ensemble & 91.5 & 96.2  \\
    DSTA~\cite{shi2020decoupled} & 4 ensemble & 91.5 & 96.4 \\
    MST-GCN~\cite{chen2021multi} & 4 ensemble & 91.5 & 96.6 \\
    EfficientGCN-B4~\cite{song2021constructing} & 4 ensemble & 91.7 & 95.7 \\
    CTR-GCN~\cite{chen2021channel} & 4 ensemble & 92.4 & 96.8 \\
    Info-GCN~\cite{Chi_2022_CVPR} & 4 ensemble & 92.7 & 96.9 \\
    \hline

\multirow{3}*{\textbf{Ours}} & Joint/Joint-M & 90.2/88.0  & 95.6/93.7 \\
& Bone/Bone-M & 91.2/87.8  & 95.5/93.2 \\
     & 4 ensemble & \textbf{92.9}  & \textbf{97.0} \\
    \bottomrule
  \end{tabular}
  }
  \caption{Action classiﬁcation performance on the NTU RGB+D dataset.}

  \label{tab:ntu}
\end{table}

\begin{table}[h]
  \centering
      \resizebox{0.45\textwidth}{35mm}{

  \begin{tabular}{lc@{}c@{}c@{}@{}}
    \toprule
    \multirow{2}*{Methods} & Mode &  \multicolumn{2}{c}{NTU-RGB+D 120} \\
         &   &  X-Sub(\%) & X-Set(\%)   \\
    \midrule
     
      SGN~\cite{zhang2020semantics} & 2 ensemble &  79.2 & 81.5 \\
    ST-TR (Plizzari et al. 2021) & 4 ensemble & 82.7 & 84.7   \\

        2S-AGCN~\cite{shi2019two} & 2 ensemble & 82.9 & 84.9 \\
    TA-CNN~\cite{xu2021topology} & 4 ensemble & 85.4 & 86.8  \\

       4S-Shift-GCN~\cite{cheng2020skeleton} & 4 ensemble & 85.9 & 87.6 \\
      DC-GCN+ADG~\cite{cheng2020decoupling} & 4 ensemble & 86.5 & 88.1 \\ 
        DSTA~\cite{shi2020decoupled} & 4 ensemble & 86.6 & 89.0   \\
        MS-G3D~\cite{liu2020disentangling} & 2 ensemble & 86.9 & 88.4  \\
      PA-ResGCN-B19~\cite{song2020stronger} & 4 ensemble & 87.3  & 88.3 \\
        Dynamic GCN~\cite{ye2020dynamic} & 4 ensemble & 87.3 & 88.6 \\
        MST-GCN~\cite{chen2021multi} & 4 ensemble & 87.5 & 88.8 \\
      EfficientGCN-B4~\cite{song2021constructing} & 4 ensemble & 88.3 & 89.1 \\

                  CTR-GCN~\cite{chen2021channel} & 4 ensemble & 88.9 & 90.6 \\
                  Info-GCN~\cite{Chi_2022_CVPR} & 4 ensemble & 89.4 & 90.7 \\

 \hline

     \multirow{3}*{\textbf{Ours}} & Joint/Joint-M & 85.5/82.3 & 87.0/83.9 \\
     & Bone/Bone-M & 87.5/82.4 & 88.7/84.4 \\
       & 4 ensemble & \textbf{89.9} & \textbf{91.1} \\
 
    \bottomrule
  \end{tabular}
  }
    \caption{
    Action classiﬁcation performance on the NTU RGB+D 120 dataset.}
  \label{tab:ntu120}
  \vspace{-3mm}

\end{table}

\subsection{Comparison with State-of-the-arts}

We compare our method with previous state-of-the-arts in Tables~\ref{tab:ucla}, \ref{tab:ntu} and \ref{tab:ntu120}. For fair comparison, we use the 4 ensembles strategy (Joint, Joint-Motion, Bone, Bone-Motion) as it is adopted by most of the previous methods. The results are means of 5 runs, the std is approximately 0.1. As shown in Table~\ref{tab:ucla}, on NW-UCLA, GAP outperforms CTR-GCN by $0.7\%$. It also outperforms the recent work Info-GCN~\cite{Chi_2022_CVPR} by $0.6\%$, which uses self-attention layer and information bottleneck. We argue that such improvement is significant considering that the model performance on this dataset is already very high. On NTU RGB+D, GAP outperforms CTR-GCN~\cite{chen2021channel} by 0.5\% on cross-subject and $0.2\%$ on cross-view settings, and it outperforms Info-GCN by $0.2\%$ and $0.1\%$ on the two settings, respectively. On the largest dataset NTU RGB+D 120, as shown in Table~\ref{tab:ntu120}, our method surpasses CTR-GCN by a large margin ($1.0\%$) on cross-subject, and $0.5\%$ on cross-set settings, respectively. Info-GCN also achieves strong performance on this dataset, while GAP still outperforms it by $0.5\%$ and $0.4\%$, respectively. In summary, GAP consistently outperforms the SOTA on NW-UCLA, NTU RGB+D and NTU RGB+D 120 under different settings, validating its effectiveness and robustness.

\subsection{Discussions}

\begin{figure}[h]
		
		\begin{center}
			\includegraphics[width=0.48\textwidth]{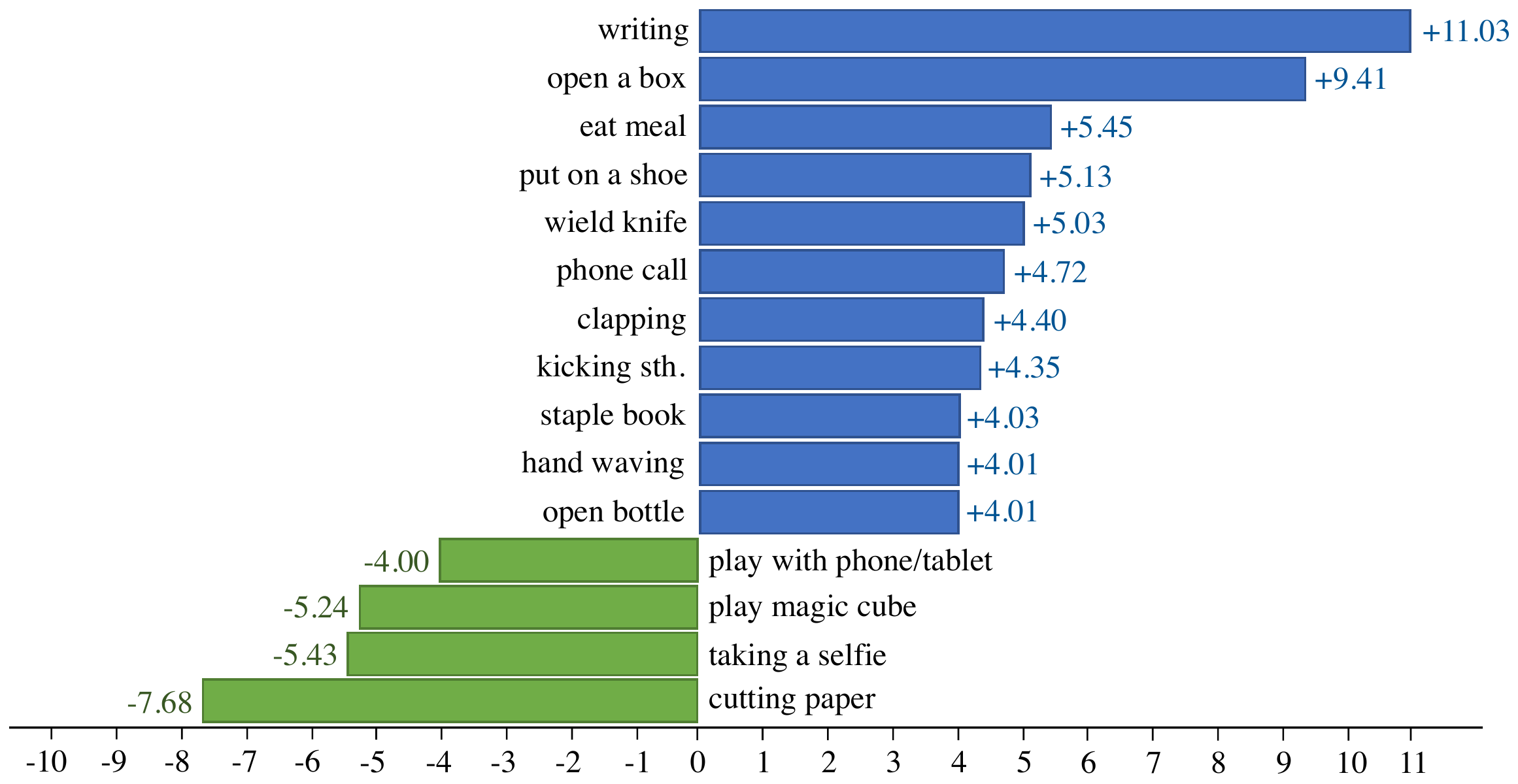}		
			\caption{Action classes with accuracy differences higher than $4\%$ between CTR-GCN and our method.}
			\label{fig:acc_analysis}
		\end{center}
		\vspace{-5mm}
	\end{figure}

To facilitate deeper discussions of the proposed GAP method, we utilized a model trained with joint modality on the NTU RGB+D 120 cross-subject mode dataset. In Figure~\ref{fig:acc_analysis}, we present the action classes that exhibit over $4\%$ absolute accuracy differences on NTU120 with and without GAP. A good case can be observed for actions such as ``writing", ``open a box," ``eat meal", and ``wield knife", which benefit significantly from GAP due to the language model generating detailed descriptions of body part movements for these actions. On the other hand, GAP performs poorly for action classes such as ``cutting paper", ``taking a selfie", ``play magic cube", and ``play with phone/tablet". Our analysis revealed that the primary distinguishing factor between these bad performing actions and good performing ones is that the former are object-related, making it challenging to recognize them using skeleton data. Additionally, the category bias present in the dataset may also contribute to the observed performance variations of our proposed method on object-related actions in NTU120 due to the presence of other action categories within the dataset. For instance, upon analyzing ``cutting paper", we found that the primary distinguishing factor between it and ``rubbing two hands" (which is also present in NTU120) is the presence of an object being held, such as paper and scissors. Conversely, although ``opening a box" is also an object-related action, there are no other object-related similar actions within the NTU120 dataset, such as ``unfold clothes". \textbf{For more discussions and visualization results, please refer to the supplementary materials.}

 
\section{Conclusion}

    We developed a novel generative action-description  prompts (GAP) framework for skeleton-based action recognition, which is the first work of its kind, as far as we known, to use action knowledge prior for skeleton action recognition. We employed large-scale language models as knowledge engine to automatically generate detailed descriptions of body parts without laborious manual annotation. GAP utilized knowledge prompting to guide skeleton encoder and enhance the learned representation with knowledge about relations of actions and human body parts. The extensive experiments demonstrated that GAP is a general framework and it can be coupled with various backbone networks to enhance representation learning. GAP achieved new state-of-the-arts on NTU RGB+D, NTU RGB+D 120 and NW-UCLA benchmarks.
    
\section{Appendix}

\subsection{Implementation details}

\textbf{Part partition.} The body part partition is slightly different in NW-UCLA as it only contains 20 joints, while NTU RGB-D and NTU RGB-D 120 contain 25 joints. However, the overall groupings remain the same, which contain four parts: head, hands, hip, legs.

\textbf{Implementation.} Our implementation is based on CTR-GCN~\cite{chen2021channel}. We also adopt the data pre-processing in InfoGCN~\cite{Chi_2022_CVPR}, where $K=1$ for bone modality and $K=8$ or $6$ for joint modality in NTU RGB-D 60/120 and NW-UCLA, respectively. For the implementation of text encoder, two different pre-training schemes (image-text and pure text) are considered. For image-text pre-training, we adopt CLIP~\cite{radford2021learning}. For pure text pre-training text encoder, we adopt Roberta\footnote{https://github.com/huggingface/transformers}. 

We implement our framework with Pytorch\footnote{https://pytorch.org}. All the models in our experiments are trained with 2 RTX 3090 GPUs and the seed is set to 1. Mixed-precision training is adopted to accelerate training speed and reduce memory footprint.

\subsection{Comparison of losses}

\begin{figure*}[h]
		\vspace{-4mm}
		\begin{center}
			\includegraphics[width=0.8\textwidth]{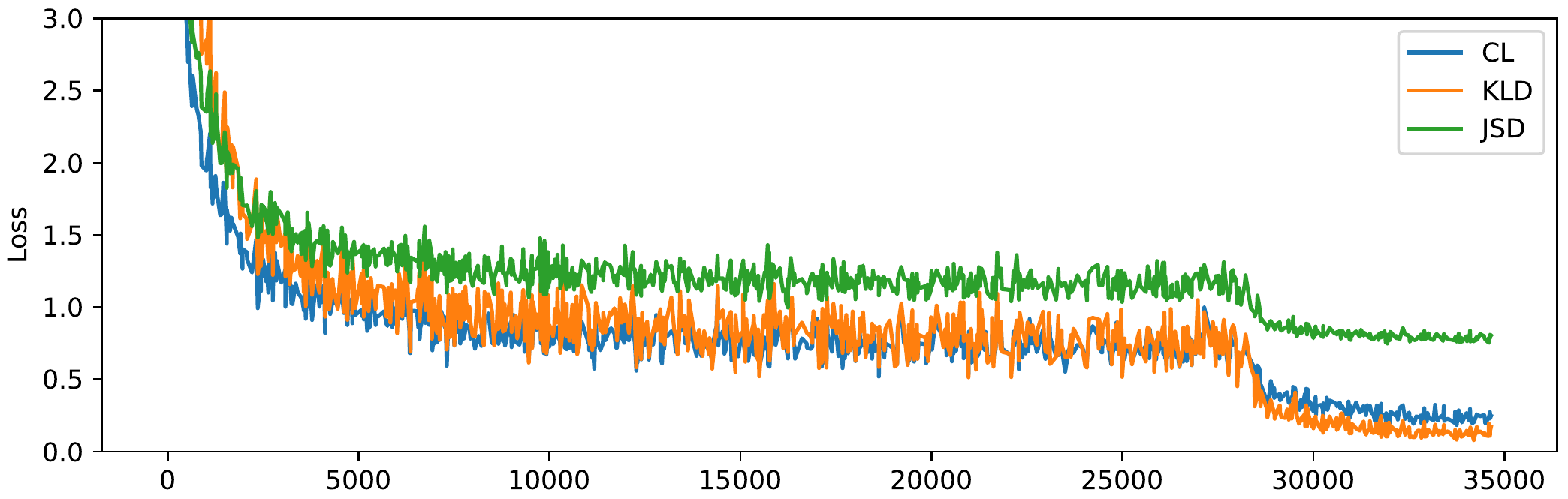}		
			\caption{Convergence curves of CL, KLD and JSD losses.}
			\label{fig:loss}
		\end{center}
		\vspace{-7mm}
	\end{figure*}

We compare 3 different losses: standard Contrastive Loss (CL), KL-Divergence (KLD) and Jenson-Shannon Divergence (JSD) on NTU120 X-sub, whose results are \textbf{85.4\%}, \textbf{85.5\%}, \textbf{85.7\%}, respectively. Fig.~\ref{fig:loss} shows that CL converges faster than KLD and JSD but with lower performance. JSD provides the best performance, probably due to its symmetric and smooth property.

\subsection{Ablation studies on ST-GCN backbone}

In Tab.~\ref{tab:stgcn}, we reran two ablation studies with weaker ST-GCN backbone to show clearer trends of our design choices. We also ran NTU120 Xsub joint for \textbf{5} runs. The mean$\pm$std is \textbf{85.46$\pm$0.13}. The mean is the same as our reported result in the main paper, which indicates that our method is relatively stable.

\begin{table*}[h]
\qquad \qquad \qquad \qquad \qquad \qquad 
  \begin{subtable}[c]{0.25\textwidth}
  \subcaption[]{Text prompt type}
    \resizebox{1\textwidth}{12.5mm}{
  \begin{tabular}{lc@{}}
    \hline    
    Model Strategy & Acc(\%) \\
    \hline
    ST-GCN    & 82.61$\pm$0.11 \\
    Label name & 82.79$\pm$0.13 \\
    Syn./Para. & 83.22$\pm$0.12 \\
    Body parts    & 83.54$\pm$0.13 \\
    Syn.+Body parts    & 83.84$\pm$0.14 \\
    \hline
  \end{tabular}}
  \end{subtable}
  \qquad 
\begin{subtable}[c]{0.2\textwidth}
  \subcaption[]{Description methods}
  \resizebox{1\textwidth}{11mm}{
  \begin{tabular}{lc@{}}
    \hline    
    Methods & Acc(\%) \\
    \hline
    Part CLS & 82.50$\pm$0.11 \\
    Manual & 83.18$\pm$0.10  \\
    HAKE & 83.21$\pm$0.15  \\
    GPT-3 & 83.84$\pm$0.14 \\
    \hline
  \end{tabular}}
  \end{subtable}
  \caption{Ablation with weaker ST-GCN backbone.}
\label{tab:stgcn}
\end{table*}

\subsection{Visualization}

\begin{figure*}[h]
		
		\begin{center}
			\includegraphics[width=1\textwidth]{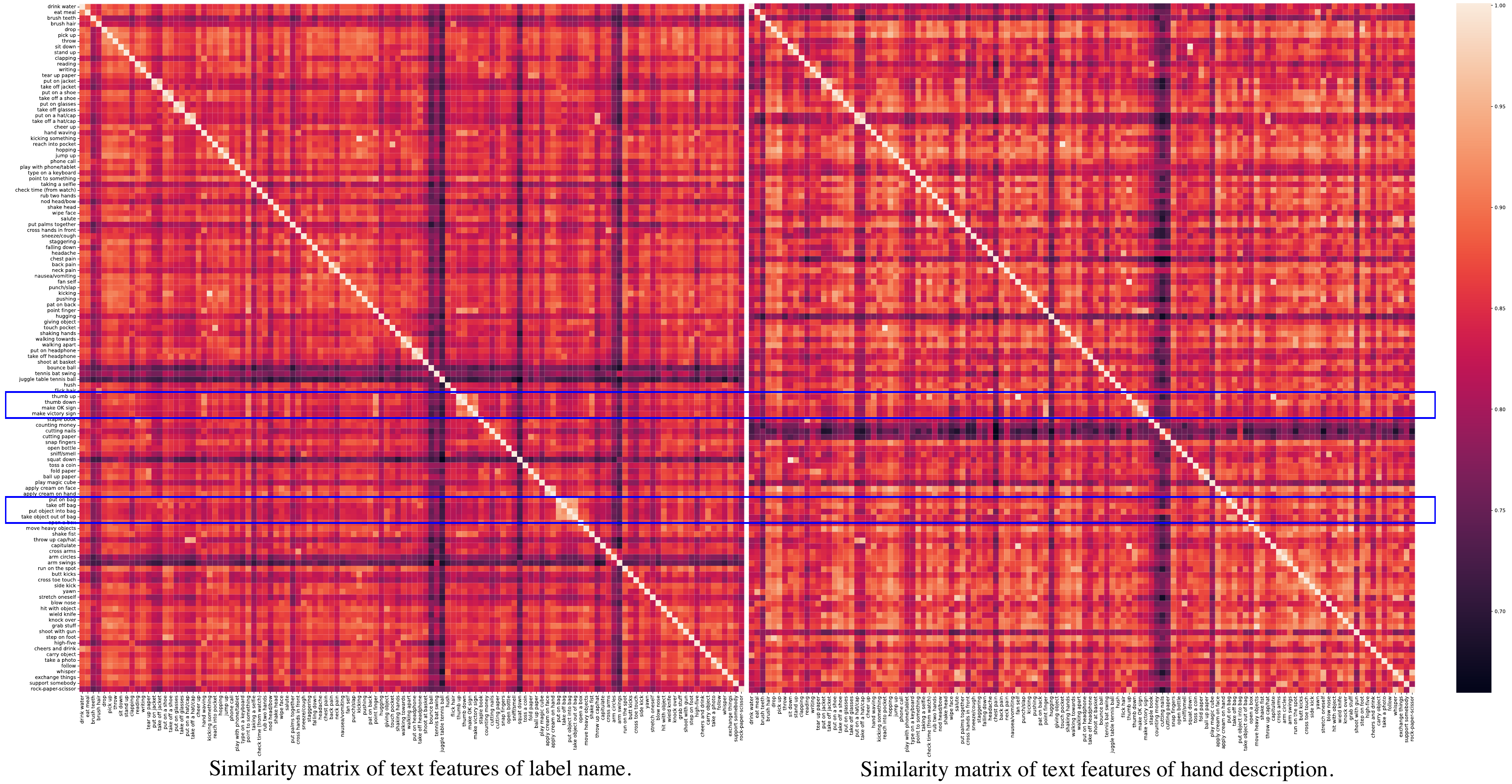}		
			\caption{Similarity matrices of text features using label name and hand part description.}
			\vspace{-4mm}
			\label{fig:label_hand}
		\end{center}
		
	\end{figure*}

\begin{figure*}[h]
		
		\begin{center}
			\includegraphics[width=1\textwidth]{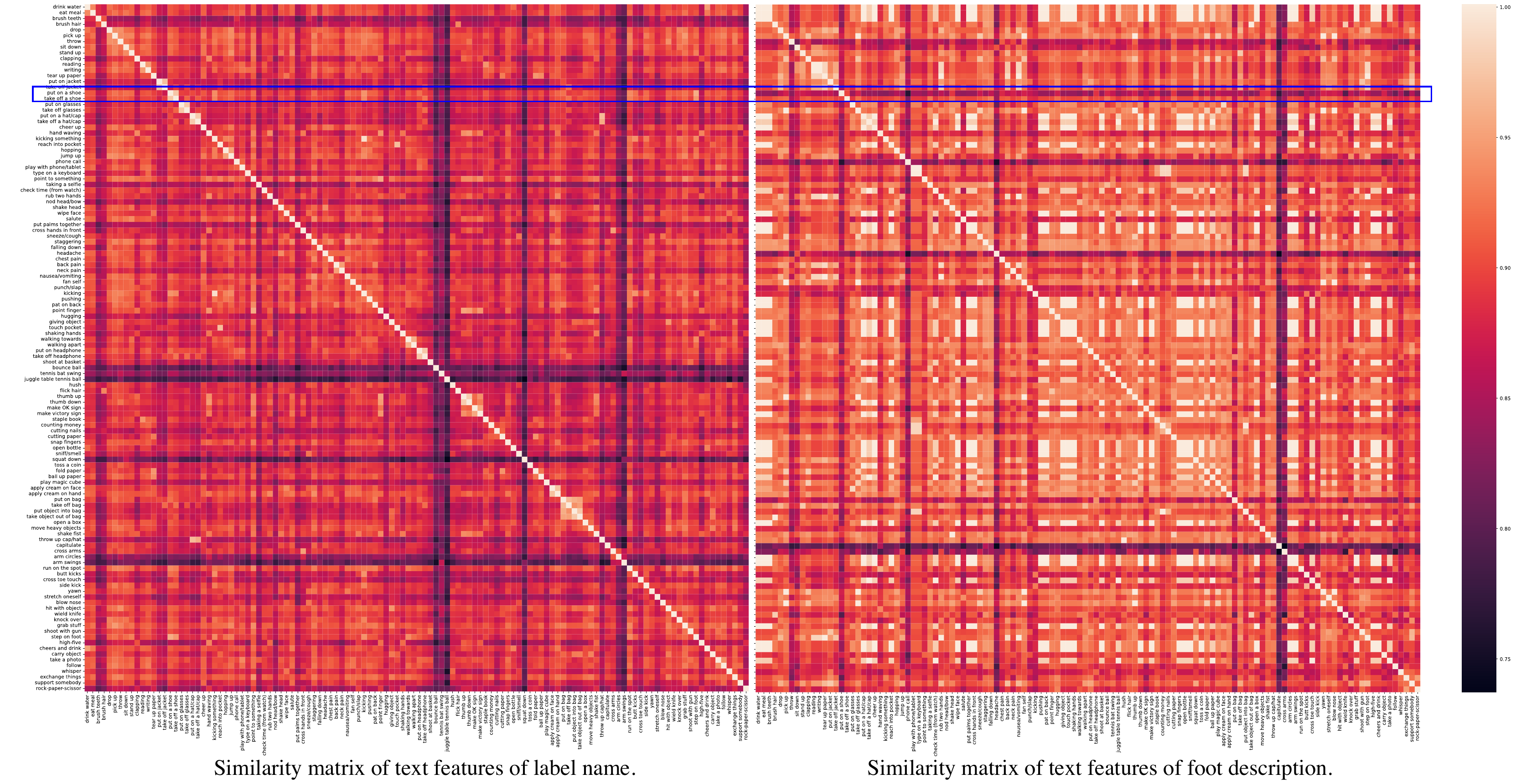}		
			\caption{Similarity matrices of text features using label name and foot part description.}
			\label{fig:label_foot}
		\end{center}
		
	\end{figure*}

In Figure \ref{fig:label_hand}, we display similarity matrices of text features derived from both the label name and the description of hand parts. It becomes evident that the descriptions of hand parts demonstrate a greater discriminative capacity for actions that primarily involve the hands, such as ``thumb up", ``thumb down", ``make OK sign", and ``make victory sign". Furthermore, these descriptions are more effective for actions centered on arm movements, such as ``put on bag", ``take off bag", ``put object into bag", and ``take object out of bag". Analogously, as illustrated in Figure \ref{fig:label_foot}, text features from foot descriptions are more applicable for actions that predominantly focus on the feet, such as ``put on a shoe" and ``take off a shoe". However, foot descriptions perform poorly on actions that primarily involve the hands. Consequently, the best results are achieved by combining these descriptions, as demonstrated in our paper.

\subsection{Text descriptions}

We provide text descriptions generated by GPT-3 (text-davinci-002) for different text prompts on NTU RGB-D 120 on \url{https://github.com/MartinXM/GAP} in \textit{paragraph-GPT3.txt}, \textit{synonym-GPT3.txt} and \textit{part-GPT3.txt}, respectively. The text prompts used for generating text descriptions are as follows: 

\begin{itemize}
    \item Describe a person [action] in details.
    \item Suggest 10 synonyms for [action].
    \item Describing following body parts actions when [action]: head, hand, arm, hip, leg, foot.
\end{itemize}

The generated descriptions are only edited to correct format error of GPT-3 without changing the semantic contents. We found that \textbf{paragraph} provides rich descriptions but sometimes with unnecessary details. \textbf{Synonym} enriches the label name meanings with similar words. \textbf{Part description} provides the most precise description of actions.

Several text descriptions are shown below:

\vspace{+4mm}

\centering
\textbf{drink water}
\begin{itemize}
    \item \textbf{Paragraph}: The man is sitting at a table with a plate of food in front of him. He is using a fork and knife to eat the food on his plate. He is chewing each bite of food before swallowing. He is also taking breaks in between bites to drink from a glass of water.
    \item \textbf{Synonym}: dine, sup, lunch, feast, banquet, picnic, breakfast, brunch, dinner, lunch.
    \item \textbf{Part description}: head tilts slightly forward; hand brings food to mouth; arm supports hand; hip remains stationary; leg remains stationary; foot remains stationary.
\end{itemize}

\textbf{eat meal}
\begin{itemize}
    \item \textbf{Paragraph}: The man is drinking water from a glass. He is holding the glass in his hand and bringing it to his lips. He is tilting the glass back and forth, allowing the water to flow into his mouth. He is swallowing the water and then setting the glass down on the table.
    \item \textbf{Synonym}: sip, guzzle, gulp, swig, chug, quaff, swill, slug, chug down, toss back.
    \item \textbf{Part description}: head tilts back slightly; hand grasps cup; arm lifts cup to mouth; hip remains stationary; leg remains stationary; foot remains stationary.
\end{itemize}

\textbf{take off jacket}
\begin{itemize}
    \item \textbf{Paragraph}: The man is taking off his jacket. He is standing up straight and reaching his arms up above his head. His jacket is coming off easily and he is taking it off quickly. He is not having any trouble taking his jacket off.
    \item \textbf{Synonym}: remove jacket, take jacket off, divest oneself of jacket, unburden oneself of jacket, get rid of jacket, be rid of jacket, dispose of jacket, get jacket off, shed jacket, slip off jacket.
    \item \textbf{Part description}: head tilts back slightly; grabs the bottom of jacket with both hands, brings hands up the jacket; arms straighten as the jacket falls downs;hip steps out of the jacket;legs straight;feet on the ground.
\end{itemize}

\textbf{make victory sign}
\begin{itemize}
    \item \textbf{Paragraph}: He is standing with his feet apart and his arms raised in the air, making a V sign with his fingers. He has a triumphant look on his face and is clearly enjoying himself.
    \item \textbf{Synonym}: give the thumbs up, give a thumbs up, give the okay sign, give the A-okay sign, give the victory sign, give a V sign, give a peace sign, give the finger, give the bird, give the one-finger salute
    \item \textbf{Part description}: head tilts slightly forward; hand forms a V shape with the index and middle fingers; arm extends fully; hip remains stationary; leg remains stationary; foot remains stationary.
\end{itemize}

\textbf{make ok sign}
\begin{itemize}
    \item \textbf{Paragraph}: The man is making an OK sign with his hand. His thumb and index finger are touching to form a circle, and his other three fingers are extended. He may be doing this to indicate that everything is okay, or to give someone the okay sign.
    \item \textbf{Synonym}: A-OK, all right, alright, copacetic, dandy, fine, good, hunky-dory, okay, swell
    \item \textbf{Part description}: head tilts slightly to the side; hand forms a circle with the thumb and first two fingers; arm extends straight out from the shoulder; hip remains at a neutral position; leg remains at a neutral position; foot remains at a neutral position.
\end{itemize}

\textbf{move heavy objects}
\begin{itemize}
    \item \textbf{Paragraph}: The man is muscular with broad shoulders. He has a deep tan, and his arms are covered in a light sheen of sweat. He grunts with effort as he lifts a heavy object, his biceps straining. His face is set in a determined expression, and his eyes are focused on the task at hand. He moves with purpose, each step sure and confident. He makes the task look easy, despite the obvious strain it is causing him.
    \item \textbf{Synonym}: drag, haul, lug, tow, carry, convey, transport, shift, relocate, remove
    \item \textbf{Part description}: head turns to look at the object; hand reaches out and grasps the object; arm pulls the object towards the body; hip moves forward to create leverage; leg pushes against the ground to create power; foot stabilizes the body.
\end{itemize}

\textbf{support somebody}
\begin{itemize}
    \item \textbf{Paragraph}: He is standing behind the person in front of him, with his hands on their shoulders. He is looking down at them with a concerned expression on his face. He is speaking quietly to them, offering words of encouragement.
    \item \textbf{Synonym}: help, assist, back, prop, buttress, shore up, strengthen, reinforce, hold up, lift up
    \item \textbf{Part description}: head tilts slightly forward; hand grasps the other person's arm just above the elbow; arm supports the other person's arm; hip stands upright; leg stands upright; foot stands flat on the ground.
\end{itemize}

\textbf{open a box}
\begin{itemize}
    \item \textbf{Paragraph}: The man is standing in front of a box. He is reaching for the lid of the box. He is opening the lid of the box. He is looking inside the box.
    \item \textbf{Synonym}: unpack, unseal, unbox, decant, disentangle, extricate, liberate, release, remove, untie
    \item \textbf{Part description}: head tilts slightly forward; hand reaches out and grasps the edge of the lid; arm extends forward; hip remains stationary; leg remains stationary; foot remains stationary.
\end{itemize}

\textbf{open bottle}
\begin{itemize}
    \item \textbf{Paragraph}: He unscrews the cap of the bottle with one hand, while holding the base of the bottle in the other. He twists the cap until it comes off with a small pop. He brings the bottle to his lips and takes a long drink, savoring the flavor of the liquid inside.
    \item \textbf{Synonym}: jar, can, container, vessel, receptacle, flask, decanter, urn, cruet, carafe
    \item \textbf{Part description}: head tilts back slightly; hand grasps the neck of the bottle; arm extends the arm holding the bottle; hip remains stationary; leg remains stationary; foot remains stationary.
\end{itemize}

\textbf{jump up}
\begin{itemize}
    \item \textbf{Paragraph}: He is a man who is of average height and build. He has dark hair and eyes, and is wearing a pair of jeans and a t-shirt. He is jump up in the air, and his arms and legs are outstretched. He has a look of concentration on his face, and he is landing on his feet.
    \item \textbf{Synonym}: leap, bound, spring, vault, hop, skip, caper, gambol, frisk, frolic
    \item \textbf{Part description}: head tilts back and the chin points up; hands come up to the chest; arms bend at the elbows and the forearms come up; hips push forward and the legs bend at the knees; legs push off the ground and the feet come up; feet land on the ground and the legs bend at the knees.

\end{itemize}

{\small
\bibliographystyle{ieee_fullname}
\bibliography{egbib}
}

\end{document}